\documentclass{article}
\usepackage{iclr2027_conference,times}

\usepackage{amsmath,amsfonts,bm}

\def\eqref#1{equation~\ref{#1}}

\def\1{\bm{1}}

\DeclareMathAlphabet{\mathsfit}{\encodingdefault}{\sfdefault}{m}{sl}
\SetMathAlphabet{\mathsfit}{bold}{\encodingdefault}{\sfdefault}{bx}{n}

\usepackage{hyperref}
\usepackage{url}
\usepackage{booktabs}
\usepackage{multirow}
\usepackage[table]{xcolor}
\usepackage{graphicx}
\usepackage{tikz}

\definecolor{iccvblue}{rgb}{0.21,0.49,0.74}
\definecolor{myred}{HTML}{A33B3B}
\definecolor{ourpurple}{HTML}{7656A6}
\hypersetup{
    colorlinks=true,
    citecolor=iccvblue,
    linkcolor=myred,
    urlcolor=ourpurple
}
\usepackage{listings}
\usepackage[most]{tcolorbox}
\tcbuselibrary{listings}

\definecolor{codebg}{HTML}{F7F7F7}       
\definecolor{codeframe}{HTML}{D0D0D0}    
\definecolor{codetext}{HTML}{303030}     
\definecolor{codekeyword}{HTML}{202020}  
\newtcblisting{psacode}{
    listing only,
    listing engine=listings,
    width=\linewidth,
    colback=codebg,
    colframe=codeframe,
    boxrule=0.5pt,
    arc=2mm,
    outer arc=2mm,
    boxsep=0pt,
    left=8pt,
    right=8pt,
    top=7pt,
    bottom=7pt,
    before skip=8pt,
    after skip=8pt,
    listing options={
        language=Python,
        basicstyle=\ttfamily\small\color{codetext},
        keywordstyle=\bfseries\color{codekeyword},
        columns=fullflexible,
        keepspaces=true,
        showstringspaces=false,
        breaklines=true,
        breakatwhitespace=true,
        tabsize=4,
        aboveskip=0pt,
        belowskip=0pt
    }
}

\title{Programmatic Search Agents: Extending Agentic Search Beyond Query Reformulation}

\author{%
  \normalfont
  \parbox[t]{\dimexpr\textwidth-2\tabcolsep\relax}{%
    \centering\normalfont
    Jiaming Qian$^{1,2,*}$ \quad
    Huiyan Yang$^{2}$ \quad
    Mandi Liu$^{2}$ \quad
    Jie Liu$^{2}$ \quad
    Wenkai Shen$^{1}$\\[4pt]
    Pengyang Zhou$^{1}$ \quad
    Jing Jin$^{3}$ \quad
    Jin Ma$^{2}$ \quad
    Dezhi Ye$^{2}$ \quad
    Chaochao Chen$^{1}$\\[6pt]
    $^{1}$ Zhejiang University \quad
    $^{2}$ Yuanbao Team, Tencent \quad
    $^{3}$ Peking University
  }%
}

\iclrfinalcopy 
\begin{document}

\maketitle
\lhead{%
  \raisebox{-2pt}{\includegraphics[height=13pt]{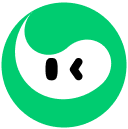}}%
  \hspace{6pt}\normalfont\normalsize\sffamily\bfseries\itshape Tencent Yuanbao
}
\begingroup
\renewcommand{\thefootnote}{\fnsymbol{footnote}}
\begin{NoHyper}
\footnotetext[1]{Work done during internship at Tencent Yuanbao.}
\end{NoHyper}
\endgroup

\begingroup
\setlength{\parskip}{4pt}
\setlength{\textfloatsep}{12pt plus 2pt minus 2pt}
\setlength{\intextsep}{10pt plus 2pt minus 2pt}
\setlength{\floatsep}{10pt plus 2pt minus 2pt}
\setlength{\abovecaptionskip}{6pt}
\vspace{-6pt}

\begin{abstract}
Search agents adapt their queries, yet fixed search interfaces leave candidate
processing and evidence presentation outside the agent's direct control.
Our trajectory analysis shows that supporting passages can be retrieved yet
never delivered to the agent; a same-page oracle intervention shows that
changing the returned evidence can reduce subsequent search.
We introduce Programmatic Search Agent (PSA), which makes a local executable
computation over candidates the unit of a search action. PSA unifies a
persistent candidate workspace, flexible primitive composition, and selective
evidence presentation. It incrementally generates program cells that reuse
candidates, execute dependent operations, and select what the agent inspects
next. The runtime resolves specified data dependencies within each cell,
while the agent adapts its search strategy across cells as new evidence arrives.
We compare PSA with the Query-based Agent and Tool-based Agent on InfoSeek-Eval
and BrowseComp-Plus using five policy backbones without task-specific training.
All three interfaces share the search substrate,
and the Tool-based Agent also shares PSA's primitives and persistent workspace.
Relative to the Query-based Agent, PSA improves macro-averaged task success by
4.00 and 7.56 percentage points on the two benchmarks, respectively;
within-backbone reductions in final-step tokens average 28.3\% and 33.9\%.
These results support extending agent control beyond query reformulation to
the processing and presentation of retrieved evidence.
Code will be released subject to approval.
\end{abstract}

\section{Introduction}

Effective search requires more than finding relevant pages: an agent must also
identify and use the evidence they contain. Large language model agents
interleave reasoning and retrieval, reformulating queries to follow leads,
resolve relations, and verify candidate
answers~\citep{jin2025searchr1trainingllmsreason,li2025webthinkerempoweringlargereasoning}.
Yet a fixed \texttt{search(query)} interface leaves candidate processing outside
the agent's direct control: the pipeline decides which candidates to retain,
how to rank them, and which passages to return. The agent can revise its query,
but cannot directly revisit those processing choices. This raises our central
question: \emph{Should agents only reformulate queries, or also control how
retrieved candidates are processed and presented?}

Our motivating analysis shows why this distinction matters.
In natural Query-based Agent trajectories on
BrowseComp-Plus~\citep{chen2026browsecompplus}, supporting passages can reach
the extraction stage without ever appearing in the agent's observations;
later search recovers only some of these omissions
(Section~\ref{sec:motivating_analysis}). A controlled oracle intervention
reveals the downstream consequences: for complete omissions, substituting a
supporting passage from the same retrieved page improves task success and
reduces subsequent search decisions in the analyzed cohort. The bottleneck
can therefore lie not only in finding evidence, but also in processing and
presenting evidence already retrieved. These findings motivate giving the
agent control over this part of the search process.

Exposing individual search operations as tools provides finer control, but
leaves open the appropriate unit of interaction. In our stepwise Tool-based
Agent, independent calls can be batched, whereas dependent stages require
separate model turns to consume preceding outputs. However, \emph{a data
dependency need not require a new model judgment}: once the agent has specified
how to retrieve, rerank, and extract, the runtime can execute that sequence
before returning control for further reasoning. Building on composable
retrieval pipelines~\citep{macdonald2020pyterrier}, executable agent
actions~\citep{wang2024codeact}, and tool planning~\citep{kim2024llmcompiler},
we make a local computation over candidates the unit of a search action
(Figure~\ref{fig:search-paradigms}).

Inspired by Perplexity's Search as Code
architecture~\citep{perplexity2026searchascode}, we introduce
\textbf{Programmatic Search Agent (PSA)}. Conditioned on the task, interaction
history, and a compact description of retained state, PSA incrementally
generates executable program cells. Three mechanisms organize this control:
a \emph{persistent candidate workspace} retains objects for later reuse;
\emph{flexible primitive composition} combines retrieval, filtering, reranking,
extraction, and deduplication through variables, collection operations, and
bounded control flow; and \emph{selective evidence presentation} determines
which results become the next observation. For example, a cell can filter and
re-extract from a retained pool without another query, then return selected
source notes while keeping the underlying candidates available. Each new
observation guides the next cell, so the search strategy evolves with the
evidence. A cell may contain a single operation when immediate feedback is
needed, or a dependent chain when its processing logic can already be specified.

We evaluate PSA against the Query-based Agent and Tool-based Agent on
InfoSeek-Eval and BrowseComp-Plus with five policy backbones and no
task-specific parameter training. All three interfaces share the same search
substrate; the Tool-based Agent also shares PSA's primitives and persistent
candidate workspace. PSA achieves the highest task success for all five
backbones on InfoSeek-Eval and four of five on BrowseComp-Plus, while using the
fewest final-step tokens in every backbone--dataset comparison. Relative to
the Query-based Agent, macro-averaged task success improves by 4.00 and 7.56
percentage points on the two benchmarks, respectively; within-backbone
reductions in final-step tokens average 28.3\% and 33.9\%. PSA also improves
macro-averaged success over the Tool-based Agent, which shares its primitives
and persistent workspace.

Our contributions are threefold: (i) an empirical analysis of evidence-delivery
limitations in fixed search pipelines, using natural trajectories and a
same-page oracle intervention; (ii) PSA, a search interface that unifies
persistent candidate reuse, flexible primitive composition, and selective
evidence presentation within executable cells; and (iii) a cross-model
evaluation of task success and final-step tokens, complemented by controlled
feedback and decision-budget experiments.

\begin{figure}
    \centering
    \includegraphics[width=\linewidth]{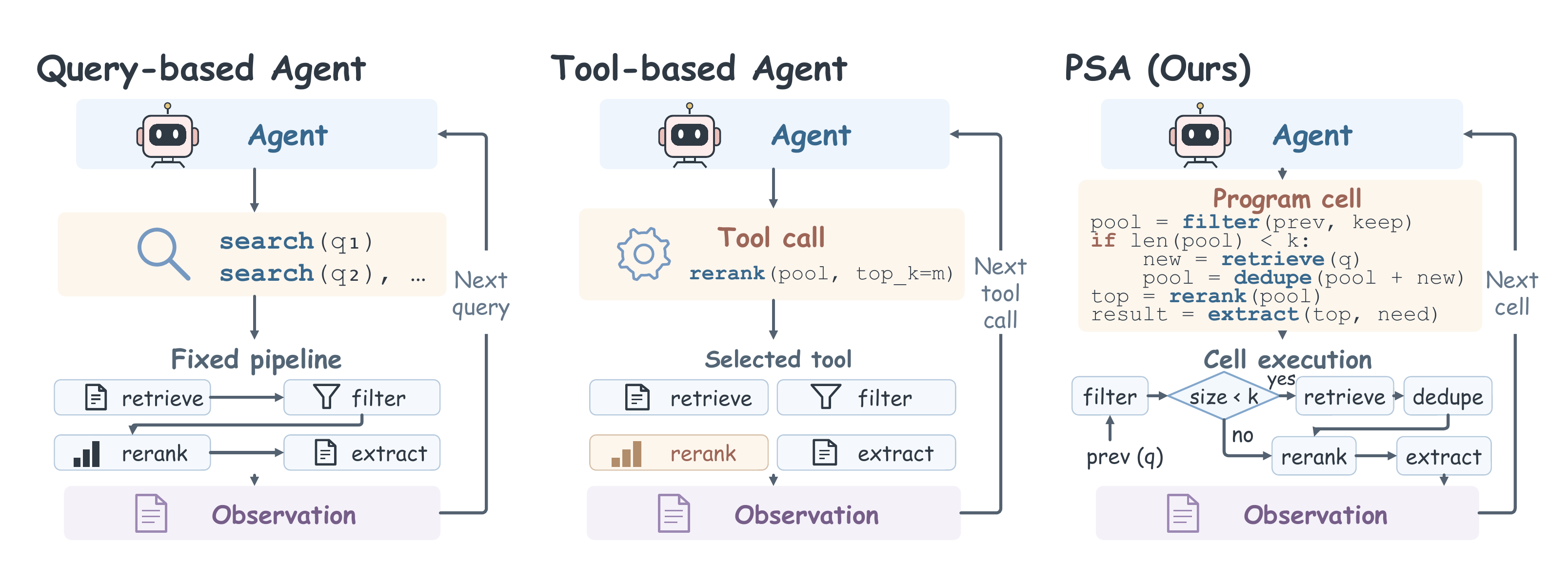}
    \caption{
    From query generation to programmatic search.
    All three agents iteratively act on search observations, but differ
    in their decision units: queries, tool calls, or a program cell.
    The Query-based Agent invokes a fixed search pipeline, the Tool-based Agent
    selects one primitive stage per turn (possibly batching independent calls),
    and PSA generates executable cells that compose operations over retained
    and new candidates. The illustrated cell filters retained candidates,
    conditionally retrieves and merges more candidates, then reranks and
    extracts evidence within one action.
    }
    \label{fig:search-paradigms}
\end{figure}

\section{The Limits of Fixed Search Pipelines}
\label{sec:motivating_analysis}

A search agent reasons over returned observations, not every candidate processed
by the search pipeline. We examine the resulting \emph{evidence-delivery gap}
in two steps: trace whether supporting passages available to extraction reach
the agent, then test how changing the returned evidence affects subsequent
search and answers while holding retrieved pages fixed at the intervention point.

\subsection{Tracing Evidence Delivery}
\label{sec:observed_delivery_gap}

\noindent\textbf{We trace supporting passages from extraction candidates to
agent observations.}
We analyze BrowseComp-Plus~\citep{chen2026browsecompplus} trajectories from the Query-based Agent using
DeepSeek-V4.1-Flash and a fixed retrieve--filter--rerank--extract pipeline.
For benchmark-annotated supporting webpages, an LLM labeling and verification
procedure identifies passages that support the reference answer or a factual
constraint in the question. A passage is \emph{available} once it enters the
extraction candidate set and \emph{delivered} when selected for a returned
search response. Cohort construction, annotation, and accounting details
appear in Appendix~\ref{app:motivation_protocol}.

\begin{figure*}[t]
    \centering
    \includegraphics[width=\textwidth]{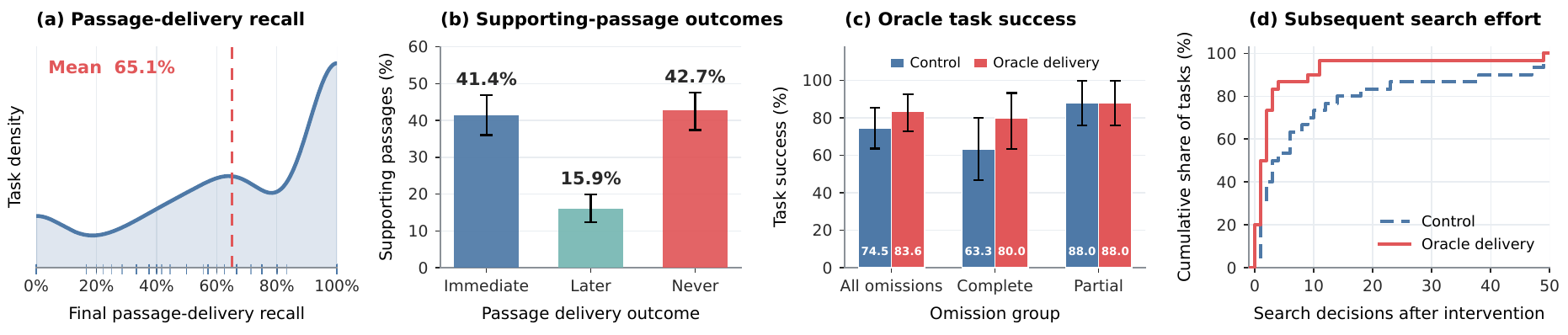}
    \caption{
    \textbf{Evidence-delivery limitations of a fixed search pipeline.}
    \textbf{(a)} Final passage-delivery recall across tasks with verified
    supporting passages (smoothed density and individual-task marks).
    \textbf{(b)} Outcomes for supporting passages: immediate delivery,
    later selection, or no delivery.
    \textbf{(c)} Judge-based task success under paired control and same-page
    oracle continuations. Error bars in (b--c) are source-stratified
    task-bootstrap 95\% confidence intervals; (c) shows per-arm uncertainty.
    \textbf{(d)} Empirical cumulative distributions of subsequent search
    decisions for complete omissions: further left indicates fewer searches,
    not higher answer accuracy. Colors follow (c). Complete/partial refers to the
    target extraction event, not the entire interaction history.
    }
    \label{fig:candidate_delivery_oracle}
\end{figure*}

\noindent\textbf{A substantial fraction of available support never reaches the
agent.}
Final passage-delivery recall ranges from no verified support delivered to
complete delivery across tasks (Figure~\ref{fig:candidate_delivery_oracle}a).
At the passage level, 42.7\% of verified supporting passages are never
delivered, although they entered the extraction candidate set; later search
recovers only a minority of the initial omissions
(Figure~\ref{fig:candidate_delivery_oracle}b). The gap therefore extends beyond
page retrieval: useful passages can be present among the candidates yet excluded
from the returned evidence.

\subsection{Intervening on Evidence Presentation}
\label{sec:delivery_oracle}

\noindent\textbf{A same-page oracle intervention tests whether omitted evidence
matters downstream.}
We replay saved tool calls to the first extraction event with verified support
and construct paired continuations when that event omits supporting passages.
From the same replayed prefix, the control receives the original response;
the oracle replaces a selected passage with omitted verified support from the
\emph{same retrieved webpage}. Both arms use the same backbone and preserve
the query, retrieved pages, pipeline, and returned item count at the
intervention point. We distinguish \emph{complete omissions}, where the target
response returns no verified support, from \emph{partial omissions}, where it
returns some but omits other supporting passages.

\noindent\textbf{Repairing complete omissions improves task success and reduces
subsequent search in this cohort.}
Oracle delivery raises judge-based task success by 16.7 percentage points and
reduces mean subsequent search decisions by 64.0\% for complete omissions
(Figure~\ref{fig:candidate_delivery_oracle}c--d). The cumulative curves show
more oracle continuations ending after only a few additional searches.
Partial omissions show no net success gain, suggesting that changing
presentation is most useful when the target response otherwise exposes no
verified support. Source-specific results and uncertainty estimates appear in
Appendix~\ref{app:motivation_protocol}.

\noindent\textbf{These findings motivate agent-controlled candidate processing.}
Query reformulation offers only indirect control over how retrieved evidence
is processed and presented. PSA makes these choices explicit through a
persistent candidate workspace, flexible primitive composition, and selective
evidence presentation. We evaluate the practical value of this control through
task success and final-step tokens under a shared search substrate.

\section{Method}
\label{sec:method}
\begin{figure}
    \centering
    \includegraphics[width=\linewidth]{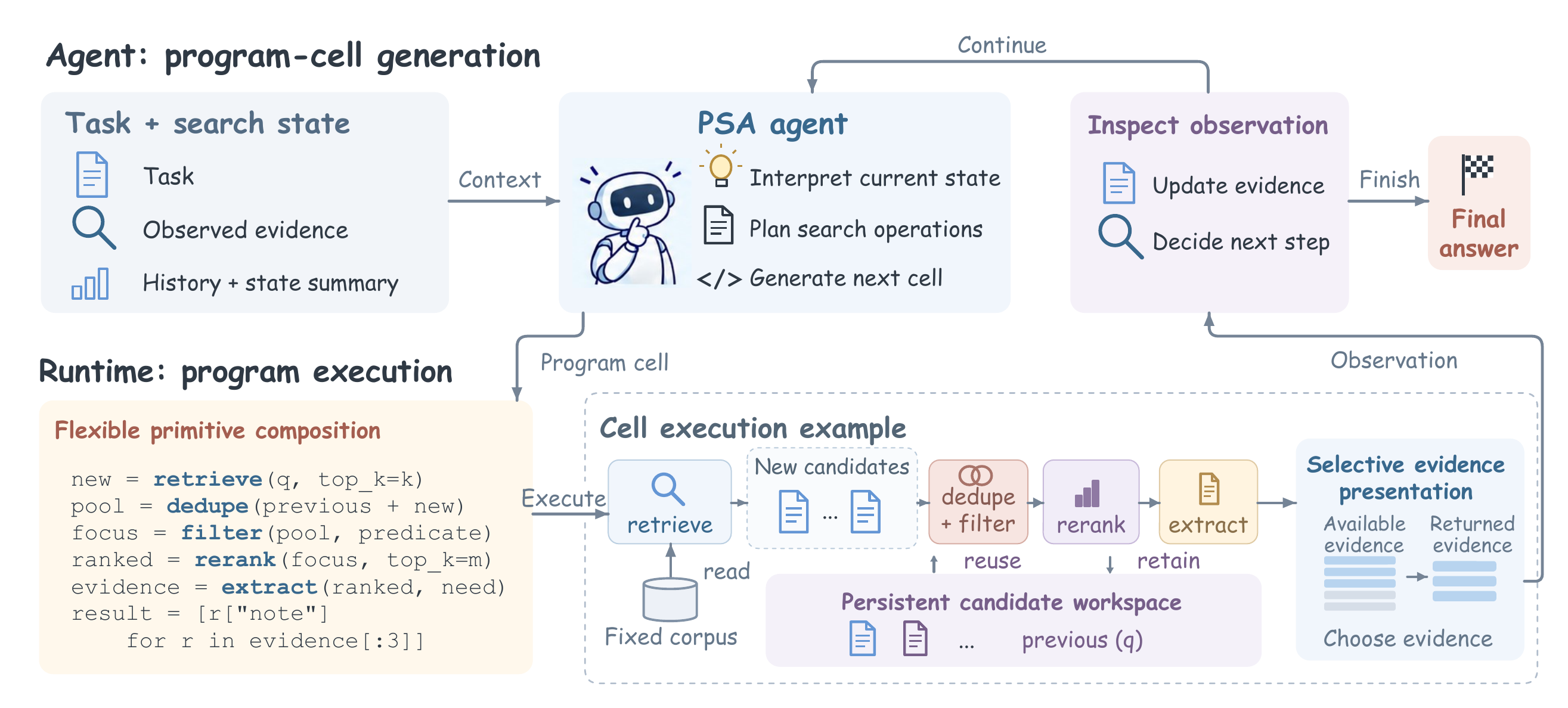}
    \caption{
    Overview of Programmatic Search Agent (PSA).
    The policy generates a program cell from the question, interaction history,
    and workspace description. A persistent workspace retains candidates;
    executable cells compose operations over them; and the selected output
    determines the next observation. The policy then chooses another cell or
    a final answer. Retained objects and returned observations are distinct.
    }
    \label{fig:psa-framework}
\end{figure}

Building on executable code actions~\citep{wang2024codeact}, PSA makes local
computation over candidates the unit of search. The policy generates an executable
\emph{program cell} that specifies search operations, their data dependencies,
and the output to inspect. The runtime executes the cell over a persistent
workspace, then returns control to the policy to interpret the result and
choose the next action (Figure~\ref{fig:psa-framework}). The interface separates
three choices: what candidate state to retain, how to process it, and what
evidence to expose for the next decision. This also separates
\emph{data dependencies}, which the runtime can resolve within a cell, from
decisions that require new model judgment. PSA uses a fixed policy backbone
without task-specific parameter updates.

\subsection{Persistent Candidate Workspace}
\label{sec:psa_state}

Candidate collections persist as addressable objects, not only as text in the
interaction history. The policy can revisit, combine, or reprocess them in
later cells without retrieving them again. It selects each action from the
history and a compact description of this retained state.
Given a question $x$ and a fixed corpus $\mathcal{D}$, its state before
interaction $t$ is
\begin{equation}
    s_t = \bigl(x, H_t, \sigma(W_t)\bigr),
    \label{eq:psa_state}
\end{equation}
where $H_t$ contains previously executed cells and their observations, and
$W_t$ stores retained candidate collections and other cell variables.
The projection $\sigma$ describes available variable names, object types, and
collection sizes, making stored objects addressable without exposing all of
their contents. Conditioned on this state, the policy selects either a new
cell or a terminal answer:
\begin{equation}
    a_t \sim \pi_{\theta}(\cdot \mid s_t),
    \qquad
    a_t \in
    \{\operatorname{Execute}(c): c \in \mathcal{C}\}
    \cup
    \{\operatorname{Finish}(y): y \in \mathcal{Y}\},
    \label{eq:psa_action}
\end{equation}
where $\mathcal{C}$ is the supported restricted-Python cell language and
$\mathcal{Y}$ the answer space. A $\operatorname{Finish}$ action terminates the
interaction; an $\operatorname{Execute}$ action updates the workspace and returns
an observation. Cells are generated incrementally, so each new observation can
change the next search target, candidate collection, or sequence of operations.
For example, a cell can store a merged evidence pool, and the next cell can
select sources from that pool for a different extraction instruction. The
source objects remain available even when their full contents were not shown
in the preceding observation.

\subsection{Flexible Primitive Composition}
\label{sec:psa_primitives}

Five shared primitives define the operations available within each cell.
Let $C$ be an ordered collection whose elements retain source identifiers,
content references, and retrieval-query provenance. Collection order determines
which candidates are selected by slicing or a per-call limit. The interfaces are
\begin{align}
    \operatorname{retrieve}(q,k)
        &\rightarrow C,
        \label{eq:retrieve}\\
    \operatorname{filter}(C,\phi)
        &\rightarrow C',
        \label{eq:filter}\\
    \operatorname{rerank}(C,k)
        &\rightarrow C',
        \label{eq:rerank}\\
    \operatorname{extract}(C,u,k)
        &\rightarrow E,
        \label{eq:extract}\\
    \operatorname{dedupe}(C_1 \Vert \cdots \Vert C_m,z)
        &\rightarrow C'.
        \label{eq:dedupe}
\end{align}
\texttt{retrieve} obtains up to $k$ candidates for query $q$;
\texttt{filter} retains those satisfying predicate $\phi$ in their original order;
and \texttt{rerank} selects up to $k$ candidates using the retrieval query stored
in their provenance. Reranking operates on a query-specific collection, keeping
its ranking objective explicit. \texttt{extract} selects up to $k$ evidence items
according to instruction $u$ and preserves their source identifiers.
In Equation~(\ref{eq:dedupe}), $\Vert$ concatenates collections, and
\texttt{dedupe} keeps the first candidate for each document or passage key $z$,
preserving survivor order. The policy supplies $q$, $u$, selection predicates,
and requested sizes; the underlying search components remain fixed
(Section~\ref{sec:experimental_setup}).

Following the compositional view of retrieval~\citep{macdonald2020pyterrier},
cells specify operation graphs rather than invoke a mandatory pipeline.
A cell may acquire new candidates, reprocess retained candidates, or combine both,
expressing dependencies and local transformations within one action.
Variable assignment connects primitive outputs to later inputs, while list
operations, bounded loops, and conditionals specify how collections are
processed. Variables can reference results from the current cell or previous
cells. Local collection operations and deduplication require no additional
retrieval or model calls. The following abbreviated excerpt from an executed cell
(Appendix~\ref{app:adaptive_cells}) reprocesses a collection retained from the
preceding cell, with no new retrieval:
\begin{psacode}
evidence_pool = merged  # Reuse the previous cell's candidates.
focus = filter(evidence_pool, predicate)
ranked = rerank(focus)
grounded = extract(ranked, instruction)
result = {"ids": [r["passage_id"] for r in grounded],
          "notes": [r["note"][:1600] for r in grounded[:3]]}
result  # Return this view to the agent; retain workspace objects.
\end{psacode}
Here, \texttt{merged} was created in the previous cell; \texttt{predicate} and
\texttt{instruction} abbreviate the original title condition and relation to
verify, given in Appendix~\ref{app:adaptive_cells}. The retained pool is filtered
before reranking and extraction, rather than reacquired through another query.
Other cells can instead retrieve and merge new evidence. Collection operations
preserve their specified order; merging pools does not implicitly rerank them.
Supported conditions and bounded loops execute on runtime values without an
intervening policy call. Decisions requiring semantic interpretation of the
returned evidence remain with the policy at the next interaction.

\subsection{Selective Evidence Presentation}
\label{sec:psa_execution}

Execution updates the workspace and returns the selected output and execution
status to the policy.
The runtime evaluates cell $c_t$ against the current workspace and fixed corpus:
\begin{equation}
    (W_{t+1}, o_t)
    =
    \mathcal{R}(c_t, W_t, \mathcal{D}),
    \label{eq:psa_execution}
\end{equation}
where $W_{t+1}$ contains the updated bindings and $o_t$ contains the selected
cell output and execution status. Execution follows the specified data
dependencies and control flow; failures are reported in the returned status.
The history is then extended as
\begin{equation}
    H_{t+1} = H_t \mathbin{\Vert} (c_t,o_t),
    \label{eq:psa_update}
\end{equation}
where $\Vert$ denotes sequence concatenation. Together, $H_{t+1}$ and
$\sigma(W_{t+1})$ form the next policy state in Equation~(\ref{eq:psa_state}).

The returned observation is a program-selected view, not a dump of the workspace.
The final variable expression chooses the value to expose: in the example,
\texttt{result} returns candidate identities and selected source notes, while
\texttt{evidence\_pool}, \texttt{focus}, and \texttt{grounded} remain addressable.
The displayed row count and note length are choices made by this cell, not
fixed interface limits. Later cells may inspect previously omitted fields or
apply a new operation to the retained objects.

The policy controls how much computation precedes each new observation.
If a cell executes $n_t$ primitive calls, they share one cell-generation
decision. A cell can also contain a single call when immediate feedback is
needed. The case in Appendix~\ref{app:adaptive_cells} illustrates all three choices:
retaining a merged pool, reprocessing it without retrieval, and exposing only
the evidence selected for the current check.

The interface comparison holds search capabilities and candidate-state access
constant between PSA and the Tool-based Agent.
The Query-based Agent invokes a fixed pipeline. The Tool-based Agent exposes
the same five primitives and persistent candidate workspace as PSA, allowing
independent calls within one primary stage (retrieve, rerank, or extract) to
be batched with filter and dedupe. Dependent primary stages require separate
replies because they consume aliases returned by preceding calls. PSA can
express those dependencies within a cell and return feedback after the
specified computation. The evaluation measures the practical value of this
interface as a whole through task success and final-step tokens.


\section{Experiments}
\label{sec:experimental_setup}

We evaluate agent control over candidate processing and evidence presentation
using task success and final-step tokens under shared search components. The main results
compare the three interfaces across policy backbones. A controlled feedback
experiment isolates the effect of exposing additional extracted evidence,
and a decision-budget sweep examines sensitivity to interaction limits.
Appendix~\ref{app:adaptive_cells} illustrates candidate reuse, primitive
composition, and selective evidence presentation in a trajectory example.
Primitive-level resource accounting is reported separately in
Appendix~\ref{app:compute_accounting}.

\subsection{Experimental Setup}

\paragraph{Datasets and evaluation.}
We use the full evaluation sets of two complementary deep-search benchmarks:
\textbf{BrowseComp-Plus} (830 questions), which pairs search-intensive questions
with a fixed webpage corpus and supporting-document annotations
\citep{chen2026browsecompplus}, and \textbf{InfoSeek-Eval} (300 questions),
the held-out InfoSeek split whose questions are synthesized from hierarchical
research trees and require resolving interdependent constraints
\citep{xia2025infoseek}. For both benchmarks, we assess answer correctness
using the respective official evaluation prompt with \texttt{GLM-5.3-Flash}
as the judge model, and report the resulting task success rate.
We report final-step tokens as the complete input messages and output tokens of
the last decision call, averaged over tasks, with the accounting convention
specified in Appendix~\ref{app:compute_accounting}.
On BrowseComp-Plus, we additionally
report Recall using its supporting-document annotations.

\paragraph{Models and search substrate.}
We evaluate all three interfaces with five policy backbones:
\texttt{DeepSeek-V4.1-Flash}, \texttt{Hunyuan-4-Preview},
\texttt{GLM-5.3-Flash}, \texttt{Qwen3.8-Flash}, and
\texttt{GPT-5.6-Luna}, using the same task set for each backbone--dataset
combination. \texttt{Hunyuan-4-Preview} uses \texttt{non\_think} mode,
and \texttt{GPT-5.6-Luna} uses \texttt{low} reasoning effort.
The policy chooses search actions and produces the final answer,
while all interfaces share a frozen search substrate: \texttt{Qwen3-Embedding-0.6B}
retrieval with an exact inner-product index, \texttt{Qwen3-Reranker-0.6B}
reranking, and deterministic passage-level extraction with
\texttt{Qwen3.5-0.8B}.  Corpora and indices are dataset-specific; component
models and primitive implementations are shared across interfaces.
GLM-5.3-Flash also serves as the answer judge. Neither policy nor
search-component models receive task-specific parameter training.

\paragraph{Interfaces and budgets.}
The Query-based Agent supports multiple parallel \texttt{search(query)} calls
per decision, each executing the fixed funnel for one query.
The Tool-based Agent exposes PSA's five primitives and persistent alias
workspace, allowing independent calls within one primary stage (retrieve,
rerank, or extract) to be batched with filter and dedupe in a turn; dependent
primary stages require separate turns to consume previously returned aliases.
PSA expresses these dependencies, local collection operations, loops, and
conditionals within one restricted-Python cell.  All interfaces use the same
runtime-enforced \texttt{20/10/10} per-call limits: retrieval returns at most
20 documents, reranking retains at most 10, and extraction selects at most
10 passages from the first 10 input candidates; smaller sets are permitted.
Each agent receives at most 50 policy decisions per task.  This budget counts
model decisions rather than primitive operations, allowing a PSA cell to
compose multiple dependent calls within one decision.
Additional runtime details and the PSA prompt are provided in
Appendix~\ref{app:runtime_details}.

\begin{table*}[t]
\centering
\small
\setlength{\tabcolsep}{2pt}
\renewcommand{\arraystretch}{1.12}

\setlength{\belowcaptionskip}{\abovecaptionskip}
\setlength{\abovecaptionskip}{0pt}
\caption{
Overall performance on the complete InfoSeek-Eval and
BrowseComp-Plus evaluation sets.
Methods within each backbone block share the search substrate and per-call
funnel.  Task success is semantically judged, and Recall measures
supporting-document recall on BrowseComp-Plus. Avg. tokens uses the final-step metric in
Section~\ref{sec:experimental_setup}; values are shown in thousands
(K = 1,000), rounded to one decimal place.
Parentheses for the Tool-based Agent and PSA give changes from the Query-based
Agent: percentage-point differences for task success and Recall, and relative
token changes computed before rounding to K ($\downarrow$ reduction;
$\uparrow$ increase). Bold marks the best value within each backbone and dataset metric.
}
\label{tab:main_results}
\resizebox{0.99\textwidth}{!}{%
\begin{tabular}{ll|cc|ccc}
\toprule
\multirow{2}{*}{\textbf{Agent Backbone}}
&
\multirow{2}{*}{\textbf{Method}}
&
\multicolumn{2}{c|}{\textbf{InfoSeek-Eval}}
&
\multicolumn{3}{c}{\textbf{BrowseComp-Plus}}
\\
\cmidrule(lr){3-4}
\cmidrule(lr){5-7}
&
&
Task success $\uparrow$
&
Avg. tokens (K) $\downarrow$
&
Task success $\uparrow$
&
Recall $\uparrow$
&
Avg. tokens (K) $\downarrow$
\\
\midrule

\multirow{3}{*}{DeepSeek-V4.1-Flash}
& Query-based Agent
& 81.00 & 33.9
& 54.94 & 59.30 & 115.7 \\
& Tool-based Agent
& 83.33\,{\scriptsize(+2.33)} & 31.3\,{\scriptsize($\downarrow$7.7\%)}
& 57.71\,{\scriptsize(+2.77)} & 62.07\,{\scriptsize(+2.77)} & 118.2\,{\scriptsize($\uparrow$2.2\%)} \\
\rowcolor[HTML]{DAEFF9}
\cellcolor{white}
& \textbf{PSA}
& \textbf{84.67}\,{\scriptsize(+3.67)} & \textbf{27.1}\,{\scriptsize($\downarrow$20.1\%)}
& \textbf{62.29}\,{\scriptsize(+7.35)} & \textbf{64.83}\,{\scriptsize(+5.53)} & \textbf{61.9}\,{\scriptsize($\downarrow$46.5\%)} \\
\midrule

\multirow{3}{*}{GLM-5.3-Flash}
& Query-based Agent & 80.33 & 15.6
& 44.94 & 40.02 & 33.2 \\
& Tool-based Agent & 81.67\,{\scriptsize(+1.34)} & 23.2\,{\scriptsize($\uparrow$49.2\%)}
& 47.11\,{\scriptsize(+2.17)} & 40.99\,{\scriptsize(+0.97)} & 37.0\,{\scriptsize($\uparrow$11.5\%)} \\
\rowcolor[HTML]{DAEFF9}
\cellcolor{white}
& \textbf{PSA}       & \textbf{82.67}\,{\scriptsize(+2.34)} & \textbf{12.9}\,{\scriptsize($\downarrow$16.9\%)}
& \textbf{51.08}\,{\scriptsize(+6.14)} & \textbf{41.09}\,{\scriptsize(+1.07)} & \textbf{20.3}\,{\scriptsize($\downarrow$38.9\%)} \\
\midrule

\multirow{3}{*}{Qwen3.8-Flash}
& Query-based Agent & 72.67 & 93.9
& 33.37 & 63.31 & 141.8 \\
& Tool-based Agent & 78.33\,{\scriptsize(+5.66)} & 73.1\,{\scriptsize($\downarrow$22.1\%)}
& 23.25\,{\scriptsize($-$10.12)} & \textbf{68.02}\,{\scriptsize(+4.71)} & 174.8\,{\scriptsize($\uparrow$23.3\%)} \\
\rowcolor[HTML]{DAEFF9}
\cellcolor{white}
& \textbf{PSA}       & \textbf{82.00}\,{\scriptsize(+9.33)} & \textbf{23.5}\,{\scriptsize($\downarrow$74.9\%)}
& \textbf{48.43}\,{\scriptsize(+15.06)} & 65.82\,{\scriptsize(+2.51)} & \textbf{74.4}\,{\scriptsize($\downarrow$47.6\%)} \\
\midrule

\multirow{3}{*}{GPT-5.6-Luna}
& Query-based Agent & 77.33 & 16.9
& 48.67 & 43.79 & 52.1 \\
& Tool-based Agent & 77.33\,{\scriptsize(0.00)} & 31.1\,{\scriptsize($\uparrow$83.4\%)}
& \textbf{52.89}\,{\scriptsize(+4.22)} & 46.66\,{\scriptsize(+2.87)} & 104.7\,{\scriptsize($\uparrow$101.0\%)} \\
\rowcolor[HTML]{DAEFF9}
\cellcolor{white}
& \textbf{PSA}       & \textbf{79.00}\,{\scriptsize(+1.67)} & \textbf{14.1}\,{\scriptsize($\downarrow$17.0\%)}
& 52.41\,{\scriptsize(+3.74)} & \textbf{47.11}\,{\scriptsize(+3.32)} & \textbf{38.1}\,{\scriptsize($\downarrow$26.8\%)} \\
\midrule

\multirow{3}{*}{Hunyuan-4-Preview}
& Query-based Agent & 76.33 & 16.6
& 57.57 & 55.63 & 52.4 \\
& Tool-based Agent & 78.00\,{\scriptsize(+1.67)} & 29.6\,{\scriptsize($\uparrow$78.6\%)}
& 59.65\,{\scriptsize(+2.08)} & 59.48\,{\scriptsize(+3.85)} & 114.3\,{\scriptsize($\uparrow$118.1\%)} \\
\rowcolor[HTML]{DAEFF9}
\cellcolor{white}
& \textbf{PSA}       & \textbf{79.33}\,{\scriptsize(+3.00)} & \textbf{14.5}\,{\scriptsize($\downarrow$12.3\%)}
& \textbf{63.06}\,{\scriptsize(+5.49)} & \textbf{65.49}\,{\scriptsize(+9.86)} & \textbf{47.3}\,{\scriptsize($\downarrow$9.7\%)} \\
\bottomrule
\end{tabular}%
}

\end{table*}

\subsection{Main Results}

\noindent\textbf{(i) PSA achieves the highest task success for all five backbones
on InfoSeek-Eval and four of five on BrowseComp-Plus.}
Table~\ref{tab:main_results} summarizes the comparison.
On InfoSeek-Eval, its macro-averaged success rate is 81.53\%, compared with
77.53\% for the Query-based Agent and 79.73\% for the Tool-based Agent, yielding gains of
4.00 and 1.80 percentage points.  On BrowseComp-Plus, the corresponding rates
are 55.45\%, 47.90\%, and 48.12\%. The BrowseComp-Plus exception is
\texttt{GPT-5.6-Luna}, for which the Tool-based Agent achieves the highest task success.

\noindent\textbf{(ii) PSA reduces final-step tokens relative to both
baselines for every backbone on both benchmarks.}
Macro-averaging the relative reductions within each backbone gives savings of
28.3\% and 46.3\% on InfoSeek-Eval, and 33.9\% and 54.5\% on BrowseComp-Plus,
relative to the Query-based Agent and Tool-based Agent, respectively. For example,
on InfoSeek-Eval with \texttt{Qwen3.8-Flash}, PSA reduces final-step
tokens from 93.9K to 23.5K relative to the Query-based Agent while increasing task
success from 72.67\% to 82.00\%.
Primitive-level accounting
for the complete DeepSeek-V4.1-Flash runs (Appendix~\ref{app:compute_accounting}) further
shows that the gains do not require uniformly increasing search-substrate
computation.

\noindent\textbf{(iii) PSA achieves higher supporting-document Recall than
the Query-based Agent across all five backbones.}
On BrowseComp-Plus, Table~\ref{tab:main_results} reports macro-averaged Recall
of 56.87\% for PSA, compared with 52.41\% for the Query-based Agent and 55.44\%
for the Tool-based Agent, corresponding to gains of 4.46 and 1.42 percentage points.
PSA has the highest reported Recall for four of five backbones; the exception
is \texttt{Qwen3.8-Flash}, where the Tool-based Agent reaches 68.02\% versus PSA's
65.82\%. The largest gain over the Query-based Agent occurs with
\texttt{Hunyuan-4-Preview}, from 55.63\% to 65.49\%.
Recall measures coverage of annotated supporting documents, complementing
answer-level success.

\subsection{Selective Evidence Presentation}
\label{sec:feedback_ablation}

\noindent\begin{minipage}[t]{0.53\textwidth}
\vspace{0pt}
\noindent\textbf{We test PSA's control over evidence presentation.}
Context management matters in long-horizon search~\citep{ye2025agentfoldlonghorizonwebagents}.
We run \texttt{GLM-5.3-Flash} once per arm on 128 BrowseComp-Plus tasks sampled
without replacement, retaining all pairs. Both arms share the PSA prompt,
feedback-format notice, persistent workspace, \texttt{20/10/10} funnel,
and 50-decision budget. Selected feedback displays the program-selected result;
full-view feedback additionally exposes all current-cell extracted rows with
complete source notes. Separate calls to the same backbone grade answers
with the official evaluation prompt.
\end{minipage}\hfill
\begin{minipage}[t]{0.44\textwidth}
\vspace{0pt}
\centering
\includegraphics[width=\linewidth]{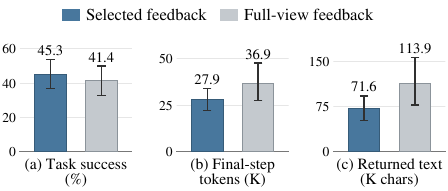}
\makeatletter\def\@captype{figure}\makeatother
\setlength{\abovecaptionskip}{3pt}
\setlength{\belowcaptionskip}{0pt}
\caption{\textbf{Evidence feedback} on 128 paired tasks,
comparing program-selected and full-view feedback.
Means and per-arm 95\% bootstrap CIs; tokens are final-step tokens.}
\label{fig:feedback_ablation}
\end{minipage}\par\medskip

\noindent\textbf{PSA's program-selected feedback reduces the evidence volume
returned to the policy.}
Compared with full-view feedback, selected feedback reduces returned text
by 37.2\% (paired bootstrap 95\% CI: 7.7--56.6\%;
Figure~\ref{fig:feedback_ablation}). The point estimates also show 3.91 percentage
points higher task success and 24.5\% lower final-step tokens (27.9K versus
36.9K). This supports separating
retention from presentation: candidates remain reusable without placing every
extracted note in the policy's context.

\subsection{Sensitivity to Decision Budgets}
\label{sec:decision_budget}

\noindent\begin{minipage}[t]{0.53\textwidth}
\vspace{0pt}
\noindent\textbf{We test PSA's quality--token trade-off across decision budgets.}
Budget-aware search calls for evaluation across interaction
limits~\citep{sun2026policybudgetinternalizingbudgetaware}.
We compare the Query-based Agent and PSA on 128 uniformly sampled BrowseComp-Plus
tasks with \texttt{Qwen3.8-Flash} at 10, 20, 30, 40, and 50 decisions.
Runs at each limit are independent, sharing the budget notice and
\texttt{20/10/10} funnel. The Query-based Agent supports parallel searches
per decision. \texttt{GLM-5.3-Flash} grades answers with the official prompt.
Appendix~\ref{app:decision_budget} gives evaluation details and uncertainty estimates.
\end{minipage}\hfill
\begin{minipage}[t]{0.44\textwidth}
\vspace{0pt}
\centering
\includegraphics[width=\linewidth]{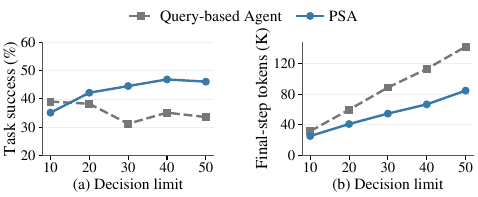}
\makeatletter\def\@captype{figure}\makeatother
\setlength{\abovecaptionskip}{3pt}
\setlength{\belowcaptionskip}{0pt}
\caption{\textbf{Decision-budget sensitivity} on all 128 sampled tasks.
Means over tasks; panel (a) spans 20--60\%.
Final-step tokens are shown in thousands (K = 1,000).}
\label{fig:decision_budget}
\end{minipage}\par\medskip

\noindent\textbf{PSA achieves higher task success with fewer final-step tokens
at decision limits of 30--50.}
In Figure~\ref{fig:decision_budget}, PSA exceeds the Query-based Agent's task success by
11.72--13.28 percentage points at these limits while using 38.3--40.9\% fewer
final-step tokens (Appendix~\ref{app:decision_budget}). PSA achieves higher
success at limits of 20--50; at 10, success is 35.16\% versus 39.06\% for
the Query-based Agent. Across the tested limits, the baseline's task success stays within
31.25--39.06\%, while its final-step token use grows from 31.7K to 141.6K.
The results support the value of agent-controlled candidate processing:
with sufficient decision budgets, PSA attains a quality--token trade-off that
additional Query-based Agent decisions do not reproduce on this cohort.

\section{Related Work}
\label{sec:related_work}

\paragraph{Iterative retrieval and agentic search.}
IRCoT and FLARE adapt retrieval to evolving reasoning and anticipated
generation~\citep{trivedi2023ircot,jiang2023active}, within the broader
reasoning--action--observation pattern of ReAct~\citep{yao2022react}.
Search-R1, ReSearch, and R-Search learn reasoning--search policies through
reinforcement learning~\citep{jin2025searchr1trainingllmsreason,chen2025researchlearningreasonsearch,zhao2025rsearchempoweringllmreasoning},
with reward design and search-engine choice affecting
training~\citep{jin2025empiricalstudyreinforcementlearning}.
WebThinker and DeepResearcher extend agentic search to web exploration and
research~\citep{li2025webthinkerempoweringlargereasoning,zheng2025deepresearcherscalingdeepresearch};
Open Data Synthesis for Deep Research and QUEST develop synthetic tasks for
training research agents~\citep{xia2025infoseek,xie2026questtrainingfrontierdeep}.
Evaluation spans browsing questions, research reports, and disentangled search
capabilities in BrowseComp, DeepResearch Bench, BrowseComp-Plus, and
DelegSearchBench~\citep{wei2025browsecompsimplechallengingbenchmark,du2025deepresearchbenchcomprehensivebenchmark,chen2026browsecompplus,yao2026delegationintelligencedeepsearch}.
PSA complements these advances by studying the search interface with fixed
backbones and shared operators, without task-specific training.

\paragraph{Candidate processing and evidence management.}
PyTerrier makes retrieval pipelines composable~\citep{macdonald2020pyterrier},
while RankGPT and pairwise ranking prompting use LLMs as ranking
operators~\citep{sun2023rankgpt,qin2024prp}. Question's Gambit pools and reranks
searches before the agent loop~\citep{rad2026questionsgambitmattersagentic}.
Beyond ranking, WebResearcher consolidates findings into reports, SLIM
summarizes long search trajectories, and Argus coordinates searchers through
an evidence graph~\citep{qiao2025webresearcherunleashingunboundedreasoning,yen2026lostmazeovercomingcontext,zhang2026argusevidenceassemblyscalable}.
AgentFold learns multi-scale folding of historical trajectories for proactive
context management~\citep{ye2025agentfoldlonghorizonwebagents}.
ACE maintains an evolving strategy playbook through generation, reflection,
and curation, using execution feedback to update context without changing
model parameters~\citep{zhang2026agenticcontextengineeringevolving}.
The Context Gathering Decision Process motivates explicit belief states and
stopping criteria; LENS localizes evidence under a budget; AnySearch learns
budget-adaptive search
policies~\citep{kausik2026contextgatheringdecisionprocess,wang2026lensincontextsearchlatent,sun2026policybudgetinternalizingbudgetaware}.
PSA focuses on agent-controlled composition of candidate-processing operations:
retained candidates can be revisited without placing every intermediate result
in context. Its motivating analysis distinguishes evidence available to the
pipeline from evidence delivered to the agent.

\paragraph{Tool use and programmatic search.}
Toolformer learns API use~\citep{schick2023toolformer}, CodeAct represents
actions as executable code~\citep{wang2024codeact}, and ReWOO and LLMCompiler
separate planning from tool
execution~\citep{xu2023rewoo,kim2024llmcompiler}.
Recursive Language Models extend programmatic execution to long-context
reasoning: the model treats input text as an external environment, inspects it
through code, and issues recursive model calls on selected
portions~\citep{zhang2026recursivelanguagemodels}.
Separating stored context from observed context is particularly relevant to
search, where candidates accumulate across queries and the evidence needed
for the next decision changes as the task unfolds.
Most directly related, Perplexity's Search as Code exposes search primitives
through an SDK that agents compose using generated Python, with control over
intermediate results~\citep{perplexity2026searchascode}.
PSA uses local executable cells over persistent candidate collections, with
new observations guiding subsequent computation. Our Tool-based Agent
shares the workspace and candidate references, allowing an interface-level
comparison under shared persistent state.

\section{Conclusion and Future Work}
\label{sec:conclusion}

Programmatic Search Agent makes executable computations over persistent
candidates the unit of search, giving the agent control over primitive
composition and evidence presentation. Persistent candidates remain available
across cells, allowing new observations to guide their reuse and processing.
Our motivating analysis shows that
fixed processing can omit useful evidence and that changing its presentation
can affect downstream outcomes. Under a shared search substrate, PSA achieves
the highest task success for all five backbones on InfoSeek-Eval and four of
five on BrowseComp-Plus, with the lowest final-step token counts for all five
on both benchmarks. PSA also improves supporting-document Recall over the
Query-based Agent for all five backbones. These results support extending
agent control beyond query reformulation to candidate processing and evidence
presentation.

A promising next step is to improve search-program generation through
trajectory-based self-distillation. Self-Distillation Policy Optimization
(SDPO) offers a relevant learning principle: use a feedback-conditioned version
of the model to supervise its policy~\citep{hubotter2026reinforcementlearningselfdistillation}.
The agent could reflect on completed
trajectories, identify effective candidate-processing strategies, and distill
them into reusable search programs. These retrospective summaries and refined
programs could serve as privileged information during training, guiding the
policy to generate effective cells from the context available at each decision.
Candidate programs could be replayed to check executability and task utility
before being used as training targets. A teacher conditioned on the retrospective
guidance could then supervise a student conditioned only on the corresponding
decision-time context.
Distilling this guidance back into the policy would aim to improve primitive
composition and evidence selection without requiring retrospective information
at inference time.

\clearpage
\endgroup

\section*{Ethics Statement}
The research conducted in this paper conforms to the ICLR Code of Ethics.

\section*{Reproducibility Statement}
Implementation and evaluation details are provided in
Section~\ref{sec:experimental_setup} and Appendix~\ref{app:runtime_details}.
We will release the code and experiment configurations to support reproduction.

\section*{LLM Usage Statement}
LLMs assisted with writing, code development, experimental analysis.

\begingroup
\hypersetup{urlcolor=black}
\bibliography{iclr2027_conference}
\bibliographystyle{iclr2027_conference}
\endgroup

\clearpage
\appendix
\begingroup
\raggedbottom
\lstdefinestyle{appendixlisting}{
    basicstyle=\ttfamily\footnotesize\color{codetext},
    keywordstyle=\bfseries\color{codekeyword},
    commentstyle=\itshape\color{codetext},
    columns=fullflexible,
    keepspaces=true,
    showstringspaces=false,
    breaklines=true,
    breakatwhitespace=true,
    tabsize=4,
    aboveskip=0pt,
    belowskip=0pt
}
\tcbset{appendixbox/.style={
    listing only,
    listing engine=listings,
    width=\linewidth,
    colback=codebg,
    colframe=codeframe,
    boxrule=0.4pt,
    arc=1mm,
    boxsep=0pt,
    left=7pt,
    right=7pt,
    top=6pt,
    bottom=6pt,
    before skip=8pt,
    after skip=8pt
}}
\newtcblisting{workspacecasecode}{
    appendixbox,
    breakable,
    pad at break*=3pt,
    listing options={style=appendixlisting,language=Python,
        basicstyle=\ttfamily\fontsize{7.5}{8.7}\selectfont\color{codetext}}
}

\section{Implementation Details}
\label{app:runtime_details}

\subsection{Search Primitives}

All interfaces share the primitive semantics in
Table~\ref{tab:primitive_contracts}. The runtime enforces the
\texttt{20/10/10} per-call funnel even though the general prompt and schemas
permit size arguments up to 50. Larger requests are capped; smaller sets are
allowed. A cell may compose several calls, so these caps and the 50-decision
budget do not equate total primitive work across interfaces.

\begin{table}[h]
\centering
\small
\renewcommand{\arraystretch}{1.12}
\setlength{\belowcaptionskip}{\abovecaptionskip}
\setlength{\abovecaptionskip}{0pt}
\caption{Shared search-primitive semantics.}
\label{tab:primitive_contracts}
\begin{tabular}{@{}p{0.17\linewidth}p{0.79\linewidth}@{}}
\toprule
\textbf{Primitive} & \textbf{Evaluation contract} \\
\midrule
\texttt{retrieve} & Returns at most 20 documents for a query. \\
\texttt{filter} & Applies field predicates with conjunction, disjunction, and negation; at most eight leaves and depth three. \\
\texttt{rerank} & Retains at most 10 documents, using their stored source query. Query branches are reranked separately before merging. \\
\texttt{extract} & Selects one passage per document from at most the first 10 input documents, preserving document order. Takes an evidence instruction and optional output size and schema. \\
\texttt{dedupe} & Removes duplicates by \texttt{doc\_id} or \texttt{passage\_id}, preserving first-seen order. \\
\bottomrule
\end{tabular}
\end{table}

\subsection{Workspace and Feedback}

\paragraph{State and observations.}
PSA submits cells through \texttt{run\_program(code)}; successful named variables
persist between cells. Candidates are read-only records retaining source
identifiers, content, and query provenance. Collection operations transform
views without modifying source records. A successful cell's final variable
expression selects its observation, while unreturned candidates remain
available. The runtime neither summarizes this output nor imposes a character limit.

\paragraph{Evidence and finalization.}
Only extraction adds to the cumulative evidence ledger. The shared fallback
finalizer receives the question and source windows from this ledger, with tools
disabled. Its system and user messages share a 16,000-character budget in the
DeepSeek-V4.1-Flash configuration, separate from the unrestricted cell-result display.

\subsection{Decision-Budget Sweep}
\label{app:decision_budget}

\paragraph{Cohort and accounting.}
The sweep in Section~\ref{sec:decision_budget} samples 128 tasks without
replacement from the 830 BrowseComp-Plus tasks.
The same cohort is used for both methods at all five
limits; no outcome-based subset is used. Each setting generates a fresh
trajectory, with temperature zero and a maximum of 4,096 output tokens per
model call. The Query-based Agent supports multiple parallel
\texttt{search(query)} calls per decision.
The tools-disabled fallback finalizer is outside the decision cap.
Judge calls are excluded from token accounting.

All 1,280 task--method--budget outcomes enter evaluation and the final-step
token means, including unsuccessful executions (one for the Query-based
Agent and 27 for PSA).

\paragraph{Uncertainty.}
We jointly resample task IDs across methods and limits for 20,000 paired
bootstrap replicates. The pointwise 95\% intervals for PSA minus Query-based
Agent task success include zero at limits 10 and 20 and lie above zero at
30--50; those for final-step tokens lie below zero at every limit.
These intervals capture task-sampling uncertainty, not repeated-run variation,
and are not adjusted for multiple comparisons.

\subsection{PSA Prompt}
\label{app:psa_prompt}

The template below condenses the experimental instructions, omitting repeated
guidance and examples. It is not the verbatim experimental prompt. The user message
contains the task question, and the native tool interface exposes
\texttt{run\_program(code)}.

\tcbinputlisting{
    appendixbox,
    breakable,
    listing file=prompts/psa_condensed.txt,
    listing options={style=appendixlisting,language={},breakindent=0pt}
}

\section{Motivation Experiment Details}
\label{app:motivation_protocol}

\subsection{Passage Annotation}

The 128-task cohort retains the original 64 tasks and adds 64 tasks sampled
without replacement from a later random 128-task cohort drawn from 830
Query-based Agent trajectories. After excluding 18 overlapping IDs, the addition
is sampled from 110 tasks, without filtering by outcome.
The combination was constructed after inspecting both runs and is a descriptive
analysis, not an independent replication or a single uniform 128-task draw.
The annotation unit is a task--passage pair from these
trajectories in Section~\ref{sec:observed_delivery_gap}. We collect passages
considered by extraction from benchmark-annotated supporting documents,
deduplicating repeated appearances within each task. Passage annotations use
\texttt{DeepSeek-V4.1-Flash} for the original cohort and
\texttt{GLM-5.3-Flash} for the added cohort, with a separate verification
pass in each. Passage labeling is separate from answer grading; the main
experiments also use \texttt{GLM-5.3-Flash} as the answer judge.

A positive passage must explicitly support the answer in the requested
relation or establish a stated question constraint. The labeler provides
an exact source quotation. Quotations absent from the passage are rejected;
the remaining positives are rechecked against the question, reference
answer, and passage in the verification pass.

\subsection{Delivery Accounting}

For each verified passage, we compare its first extraction-candidate event
with its first selection event. Equal events indicate immediate delivery;
a later selection indicates recovery; no selection indicates non-delivery.
Final passage-delivery recall is the per-task fraction of verified available
passages selected at any point. The macro-average covers the 92 tasks with
verified passages, while passage-level proportions pool their 370 passages.
The other 36 tasks are excluded from recall averages. This passage-ID accounting
does not measure semantic observation coverage. Later selection alone does not
establish query reformulation; extraction-event order is not decision order.
Confidence intervals use 20,000 task-cluster bootstrap replicates,
retaining all passages within each sampled task and resampling separately
within the two source cohorts.

Of the verified passages, 153 (41.4\%) are delivered immediately, 59 (15.9\%)
are selected later, and 158 (42.7\%) are never delivered. Later selection
recovers 27.2\% of initially omitted passages. Mean final passage-delivery recall
is 65.1\%, with seven tasks receiving none of their verified support and
30 receiving all of it.

\subsection{Same-Page Intervention}

\paragraph{Paired construction.}
We identify the first extraction event at which a verified passage becomes
available and retain tasks with an omission at that event. For each affected
document, the oracle replaces the selected passage with an omitted verified
passage from the same document, breaking ties by passage-ID order. The
replacement preserves one excerpt per document and the total number of
items. There are 55 paired tasks (31 original and 24 added). The query, retrieved documents,
and pipeline are fixed at the intervention point; both arms then continue freely.
The cumulative evidence ledger retains original entries and adds the substituted
evidence, so its total entry count is not held fixed.

Saved tool calls reconstruct the prefix; unrecorded assistant reasoning is not
reconstructed. Before substitution, rendered observations must match the source
trajectory exactly. Both arms continue with the same DeepSeek-V4.1-Flash
backbone as the source runs in both the original and added cohorts. This is a comparison
of matched continuations, not of the new oracle against the original completed
trajectory. Both cohorts use a 50-decision total cap and 4,096 output tokens per
call. Settings are matched within each pair, but action-repair limits differ
between source runs (zero in the original cohort, two in the added cohort).

\paragraph{Evidence-omission groups.}
A complete omission delivers none of the verified passages available at the
target event; a partial omission delivers some but omits others. The cohort
contains 30 complete-omission and 25 partial-omission pairs. This event-local definition
does not exclude support in preceding observations or other calls in the same
decision. All 55 oracle interventions
were applied, with no prefix-observation mismatches in either arm.
All pairs are retained, including protocol failures and decision-limit terminations.

\paragraph{Metrics and uncertainty.}
Task success uses the official BrowseComp answer-grading prompt with
GLM-5.3-Flash for both arms and source cohorts. Subsequent search
counts distinct decisions containing a search call strictly after the target
decision. The ECDF includes all terminations, including failures, and is not a
success-within-budget curve. Effect intervals use 20,000 paired-task bootstrap
replicates stratified by source.
Across all 55 pairs, mean search decisions fall from 9.71 to 6.40; the paired
difference is $-3.31$ (95\% CI: $[-7.07,0.36]$).
Figure~\ref{fig:candidate_delivery_oracle}c shows per-arm task-bootstrap
intervals, not intervals of paired differences.

Overall task success is 74.5\% versus 83.6\% (control versus oracle), with
a paired difference of $+9.1$ percentage points (95\% CI: $[-1.8,20.0]$).
For complete omissions, success is 63.3\% versus 80.0\%, a difference of
$+16.7$ points (95\% CI: $[3.3,33.3]$), while mean subsequent search decisions
are 10.73 versus 3.87, a difference of $-6.87$ (95\% CI: $[-11.73,-2.87]$).
Search counts decrease for 17 tasks, remain unchanged for nine, and increase
for four. Partial omissions have 88.0\% success in both arms.

\paragraph{Source-specific results.}
The original cohort contributes 19 complete-omission pairs, with task success
57.9\% versus 84.2\% and search decisions 11.05 versus 2.84 (control versus
oracle). The added cohort contributes 11 complete-omission pairs, with task
success 72.7\% in both arms and search decisions 10.18 versus 5.64.
Thus, the pooled success gain comes from the original cohort, while both
cohorts show lower mean search counts for complete omissions. The combined
confidence intervals do not account for the post-hoc cohort construction or
eliminate between-run annotation and configuration differences.

\section{Search-Compute Accounting}
\label{app:compute_accounting}

Table~\ref{tab:compute_accounting} reports per-task resource use from the
complete DeepSeek-V4.1-Flash traces, with the Query-based Agent as the
fixed-pipeline reference under shared components and per-call limits.
Final-step tokens comprise the complete input messages and output tokens of
the last decision call, averaged over tasks.
For historical traces without per-call usage, calibrated reconstructions are
used; reported values additionally include independent fallback-finalization
usage when present and exclude unestimable tasks from the averages.

\begin{table}[h]
\centering
\small
\setlength{\tabcolsep}{4pt}
\renewcommand{\arraystretch}{1.15}
\setlength{\belowcaptionskip}{\abovecaptionskip}
\setlength{\abovecaptionskip}{0pt}
\caption{Mean per-task resources in the complete DeepSeek-V4.1-Flash runs.
Pooled means weight the 300 InfoSeek-Eval and 830
BrowseComp-Plus tasks equally. Rerank pairs count scored documents; extract
pairs count passages scored by the passage locator. Final-step tokens are shown in thousands.}
\label{tab:compute_accounting}
\begin{tabular*}{\linewidth}{@{\extracolsep{\fill}}lrrrrrr@{}}
\toprule
\multirow{3}{*}{\textbf{Resource}}
& \multicolumn{2}{c}{\textbf{InfoSeek-Eval}}
& \multicolumn{2}{c}{\textbf{BrowseComp-Plus}}
& \multicolumn{2}{c}{\textbf{Pooled}} \\
\cmidrule(lr){2-3}\cmidrule(lr){4-5}\cmidrule(lr){6-7}
& \multicolumn{1}{c}{Query-based}
& \multicolumn{1}{c}{\multirow{2}{*}{PSA}}
& \multicolumn{1}{c}{Query-based}
& \multicolumn{1}{c}{\multirow{2}{*}{PSA}}
& \multicolumn{1}{c}{Query-based}
& \multicolumn{1}{c}{\multirow{2}{*}{PSA}} \\
& \multicolumn{1}{c}{Agent} &
& \multicolumn{1}{c}{Agent} &
& \multicolumn{1}{c}{Agent} & \\
\midrule
Retrieval requests & 9.12 & 9.79 & 43.52 & 30.88 & 34.39 & 25.28 \\
Returned documents & 182.5 & 195.8 & 870.5 & 617.6 & 687.8 & 505.6 \\
Rerank pairs       & 178.2 & 195.7 & 870.5 & 617.5 & 686.7 & 505.5 \\
Extract calls      & 9.12 & 9.79 & 43.52 & 30.84 & 34.39 & 25.25 \\
Extract pairs      & 195.4 & 228.7 & 12,729.2 & 8,437.9 & 9,401.6 & 6,258.4 \\
Final-step tokens (K) & 33.9 & 27.1 & 115.7 & 61.9 & 94.0 & 52.7 \\
\bottomrule
\end{tabular*}
\end{table}

\paragraph{Resource trade-offs.}
PSA reduces retrieval, reranking, and extraction work on BrowseComp-Plus and
in the pooled results; on InfoSeek-Eval, these counts increase while task
success improves and final-step tokens decrease.

\section{Candidate Reuse and Primitive Composition: A Case Study}
\label{app:adaptive_cells}
\label{sec:case_study}

The question links a World War I battle to commanders of the New Zealand
Expeditionary Force (NZEF) and its World War II successor, with 372 casualties
among 600 Australian troops. \texttt{Hunyuan-4-Preview}
answers \emph{Battle of the Nek} after Cell~13; consecutive Cells~6--8 below
are an intermediate excerpt.

\noindent\textbf{Candidate reuse and composition.}
Cell~6 merges two extracted branches into 20 retained candidates.
Without retrieval, Cell~7 filters three Godley-branch sources sharing a
reranking query, reranks them, and extracts two evidence items.

\noindent\textbf{Selective presentation and adaptation.}
Cell~6 displays four detailed notes, omitting Godley's; Cell~7 exposes his
NZEF command while retaining the pool. Cell~8 searches for casualty and
successor-force evidence, illustrating adaptation across cells.
Query and extraction strings are abbreviated; dependencies, primitive order,
predicates, and observation projections are preserved.

\begin{workspacecasecode}
# Cell 6: collect two branches and retain their merged evidence.
hits = retrieve("Sari Bair Godley New Zealand Division Nek casualties")
ranked = rerank(hits)
left_grounded = extract(ranked, "operation commander and Nek casualties")
bridge_hits = retrieve("Alexander Godley commander NZEF First World War")
ranked = rerank(bridge_hits)
right_grounded = extract(ranked, "Godley's NZEF command and battles")
merged = dedupe(left_grounded + right_grounded, key="passage_id")
result = {
    "frontier": [{"passage_id": r["passage_id"], "title": r["title"],
                  "value": r["value"]} for r in merged],
    "details": [{"passage_id": r["passage_id"], "note": r["note"][:1200]}
                for r in merged[:4]]}
result
# Retained: 20 candidates. Returned: 20 short entries, 4 detailed notes.

# Cell 7: reprocess the retained pool; no retrieve call.
evidence_pool = merged
focus = filter(evidence_pool, {"field": "title", "op": "in", "value": [
    "Alexander Godley", "New Zealand Expeditionary Force",
    "Military history of New Zealand during World War I"]})
ranked = rerank(focus)
grounded = extract(ranked, "Godley's WWI command and battles; Freyberg's WWII command")
result = {
    "frontier": [{"passage_id": r["passage_id"], "title": r["title"],
                  "value": r["value"]} for r in grounded],
    "details": [{"passage_id": r["passage_id"], "note": r["note"][:1600]}
                for r in grounded[:3]]}
result
# Actual flow: 20 retained candidates -> filter 3 -> rerank 3 -> extract 2.
# Returned Godley evidence supports his NZEF command throughout WWI.

# Cell 8: acquire evidence for the remaining relations.
hits = retrieve("Battle of the Nek")
ranked = rerank(hits)
left_grounded = extract(ranked, "Australian troop count, casualties, commander")
bridge_hits = retrieve("2nd NZEF Bernard Freyberg WWII successor to NZEF")
ranked = rerank(bridge_hits)
right_grounded = extract(ranked, "commander of the 2nd NZEF in WWII")
merged = dedupe(left_grounded + right_grounded, key="passage_id")
result = {
    "frontier": [{"passage_id": r["passage_id"], "title": r["title"],
                  "value": r["value"]} for r in merged],
    "details": [{"passage_id": r["passage_id"], "note": r["note"][:1500]}
                for r in merged[:4]]}
result
\end{workspacecasecode}

\endgroup

\end{document}